\documentclass{article}
\PassOptionsToPackage{authoryear, compress}{natbib}
\usepackage[preprint,nonatbib]{neurips_2019}

\usepackage[utf8]{inputenc} 
\usepackage[T1]{fontenc}    
\usepackage{url}            
\usepackage{booktabs}       
\usepackage{amsfonts}       
\usepackage{nicefrac}       
\usepackage{microtype}      

\usepackage[utf8]{inputenc}   
\usepackage[english]{babel} 
\addto\captionsenglish{}
\addto\captionsenglish{}
\addto\captionsenglish{}

\addto\captionsportuguese{}

\usepackage{graphicx}

\usepackage{mathtools}
\usepackage{amsthm}   
\usepackage{amsfonts} %

\usepackage{subfigure}

\usepackage{subfigmat}

\usepackage{dcolumn}
\newcolumntype{d}{D{.}{.}{-1}} 
\newcolumntype{e}{D{E}{E}{-1}} 

\usepackage{nomencl}
\makenomenclature
\RequirePackage{ifthen} 
\ifthenelse{\equal{\languagename}{english}}%
    { 
    \renewcommand{\nomgroup}[1]{%
      \ifthenelse{\equal{#1}{R}}{%
        \item[\textbf{Roman symbols}]}{%
        \ifthenelse{\equal{#1}{G}}{%
          \item[\textbf{Greek symbols}]}{%
          \ifthenelse{\equal{#1}{S}}{%
            \item[\textbf{Subscripts}]}{%
            \ifthenelse{\equal{#1}{T}}{%
              \item[\textbf{Superscripts}]}{}}}}}%
    }{
    \renewcommand{\nomgroup}[1]{%
      \ifthenelse{\equal{#1}{R}}{%
        \item[\textbf{Simbolos romanos}]}{%
        \ifthenelse{\equal{#1}{G}}{%
          \item[\textbf{Simbolos gregos}]}{%
          \ifthenelse{\equal{#1}{S}}{%
            \item[\textbf{Subscritos}]}{%
            \ifthenelse{\equal{#1}{T}}{%
              \item[\textbf{Sobrescritos}]}{}}}}}%
    }%

\usepackage{rotating}
\usepackage[pdftex]{hyperref} 
\hypersetup{bookmarks=true,         
            bookmarksopen=false,    
           bookmarksnumbered=true, 
            pdftitle={Variational Mixture of Normalizing Flows},
            pdfauthor={Guilherme Pires},
            pdfsubject={VMoNF},
            pdfstartview=FitV,
            pdfdisplaydoctitle=true}

\usepackage[figure,table]{hypcap}

\usepackage[
    backend=biber,
    style=authoryear,
    sortlocale=en_EN,
    natbib=true,
    url=false,
    doi=true,
    eprint=false
]{biblatex}

\newrobustcmd*{\parentexttrack}[1]{%
  \begingroup
  \blx@blxinit
  \blx@setsfcodes
\blx@bibopenparen#1\blx@bibcloseparen
 \endgroup}

\AtEveryCite{%
  }

\makeatother

\usepackage{notoccite}

\usepackage{multirow}

\usepackage{booktabs}

\usepackage{pdfpages}

\usepackage{enumitem}
\usepackage{arydshln}
\usepackage{tikz}
\setlist{nosep}

\def\bm#1{\mathchoice                             
  {\mbox{\boldmath$\displaystyle#1$}}%
  {\mbox{\boldmath$#1$}}%
  {\mbox{\boldmath$\scriptstyle#1$}}%
  {\mbox{\boldmath$\scriptscriptstyle#1$}}}

\usepackage{tikz}
\usetikzlibrary{matrix,decorations.pathreplacing,calc}

\usepackage{bm}

\usepackage{color}
\usepackage{listings}
\usepackage{caption}

\lstnewenvironment{algorithm}[1][] 
{
    \refstepcounter{nalg} 
    \captionsetup{labelformat=algocaption,labelsep=colon} 
    \lstset{ 
        mathescape=true,
        frame=single,
        captionpos=b,
        aboveskip=25pt,
        belowskip=25pt,
        numbers=none,
        keywordstyle=\color{black}\bfseries,
        keywords={for, input, output, return, datatype, function, in, if, else, foreach, while, begin, end, } 
        xleftmargin=.04\textwidth,
        #1 
    }
}
{}

\newcommand{\q}[1]{``#1''} 

\usepackage{diagbox}

\title{Variational Mixture of Normalizing Flows}

\author{%
  Guilherme G. P. Freitas~Pires\\
  Instituto Superior Técnico, Universidade de Lisboa\\
  Lisboa, Portugal\\
  \texttt{mail@gpir.es}\\
  \And
  Mário A. T. Figueiredo\\
  Instituto de Telecomunicações, \\
  Instituto Superior Técnico, Universidade de Lisboa\\
  Lisboa, Portugal\\
  \texttt{mario.figueiredo@tecnico.ulisboa.pt}\\
}

\begin{document}

\maketitle

\begin{abstract}
In the past few years, deep generative models, such as generative adversarial networks
\autocite{GAN}, variational autoencoders \autocite{vaepaper}, and their variants,
have seen wide adoption for the task of modelling complex data distributions.
In spite of the outstanding sample quality achieved by those early methods,
they model the target distributions \emph{implicitly}, in the sense that the probability
density functions induced by them are not explicitly accessible. This fact renders those methods unfit for
tasks that require, for example, scoring new instances of data with the learned
distributions. Normalizing flows have overcome this limitation by leveraging the
change-of-variables formula for probability density functions, and by using
transformations designed to have tractable and cheaply computable Jacobians. Although
flexible, this framework lacked (until recently \autocites{semisuplearning_nflows, RAD}) a way to introduce discrete structure (such as the one found in mixtures) in the models it allows to
construct, in an unsupervised scenario. The present work overcomes this by using normalizing flows as components in a mixture model and devising an end-to-end training procedure for such a model.
This procedure is based on variational inference, and uses a variational posterior
parameterized by a neural network. As will become clear, this model naturally
lends itself to (multimodal) density estimation, semi-supervised learning, and
clustering. The proposed model is illustrated on two synthetic datasets, as well
as on a real-world dataset.

\vspace{5mm}

\end{abstract}

\section{Introduction}
\label{section:introduction}

\subsection{Motivation and Related Work}
\label{subsection:motivation}

Generative models based on neural networks -- \textit{variational autoencoders} (VAEs),
\textit{generative adversarial networks} (GANs), normalizing flows, and their variations --
have experienced increased interest and progress in their capabilities. VAEs
\autocite{vaepaper} work by leveraging the reparameterization trick to optimize
a variational posterior parameterized by a neural network, jointly with the generative
model \textit{per se} - it too a neural network, which takes samples from a latent distribution at its
input space and \emph{decodes} them into the observation space. GANs also work by
jointly optimizing two neural networks: a \emph{generator}, which learns to produce
realistic samples in order to \q{fool} the second network -- the \emph{discriminator} -- which
learns to distinguish samples produced by the generator from samples taken from real data. GANs \autocite{GAN} learn
by having the generator and discriminator \q{compete}, in a game-theoretic sense. 
Both VAEs and GANs learn \emph{implicit} distributions of the data, in the sense that - if training is
successful - it is possible to sample from the learned model, but there is no direct access to the likelihood function of the learned distribution. 

Normalizing flows \autocite{shakir_nf} differ from VAEs and GANs, among other aspects, in the fact that they allow learning \emph{explicit} distributions from data\footnote{In fact, recent work \autocite{flowgan} combines the training framework of GANs with the use of normalizing flows, so as to obtain a generator for which it is possible to compute likelihoods.}. Thus, normalizing flows lend themselves to the task of density estimation.

Less (although some) attention has been given to the extension of these types
of models with discrete structure, such as the one found in finite mixtures.
Exploiting such structure, while still being able to benefit from the expressiveness
of neural generative models -- specifically, normalizing flows -- is the goal of this
work. Specifically, this works explores a framework to learn a mixture of normalizing
flows, wherein a neural network classifier is learned jointly with the mixture
components. Doing so will naturally produce an approach which performs, not
only density estimation, but also clustering, since the classifier can be used
to assign points to clusters. Naturally, this approach also allows doing semi-supervised learning, where available labels can be used to refine the classifier and selectively train the mixture components.

The work herein presented intersects several active directions of research. In the sense of combining deep neural networks with probabilistic modelling, particularly with the goal of endowing simple probabilistic graphical models with more expressiveness, \textcite{svae} and \textcite{lin2018variational} proposed a 
framework to use neural-network-parameterized likelihoods, composed with latent probabilistic graphical models. Still in line with this topic, but with an approach more focused towards clustering and semi-supervised learning, \textcite{gmVAE} proposed a VAE-inspired model, where the prior is a Gaussian mixture. Finally, \textcite{DEC} described  an unsupervised method for clustering using deep neural networks, which is a task that can also be fulfilled by the model presented in this work.

The two prior publications that are most related to the present work are those by \textcite{RAD} and \textcite{semisuplearning_nflows}. As in this paper, \textcite{RAD} tries to reconcile normalizing flows with a multimodal (or discrete) structure. They do so by partitioning the latent space into disjoint subsets, and using a mixture model where each component has non-zero weight exclusively within its respective subset. Then, using a set-identification function and a piece-wise invertible function, a variation of the change-of-variable formula is devised. \textcite{semisuplearning_nflows} also exploit a multimodal structure, while using normalizing flows for expressiveness. However, while the present work relies on a variational posterior parameterized by a neural network and learns $K$ flows (one for each mixture component), the method proposed by \textcite{semisuplearning_nflows} resorts to a latent mixture of Gaussians as the base distribution for its flow model, and learns a single normalizing flow.

\subsection{Contributions}
\label{subsection:objectives}

The main contribution of the present work can be summarized as follows: we propose a finite mixture of normalizing flows  with a tractable end-to-end learning procedure. We also provide a proof-of-concept implementation to demonstrate the capabilities of such model, and illustrate its working in three different types of leaning tasks: density estimation, clustering, and semi-supervised learning.

We have achieved these goals by proposing a method to learn a mixture of $K$ normalizing flows, through the optimization of a variational inference objective, where the variational posterior is parameterized by a neural network with a softmax output, and its parameters are optimized jointly with those of
the mixture components.

\subsection{Notation}
\label{section:notation}
The main notation used throughout this work is as follows. Scalars and vectors are lower-case letters, with vectors in bold (\textit{e.g.}, $x$ is a scalar, $\mathbf{z}$ is a vector). Upper-case letters represent  matrices. Vector $\bm{x}_{a:b}$ contains the $a$-th to the $b$-th elements of vector $\bm{x}$. For distributions, subscript notation will only be used when the distribution is not clear from context. The operator $\odot$ denotes the element-wise product. The letter $x$ is preferred for observations. The letter $z$ is preferred for latent variables.  The letter $\bm\theta$ is preferred for parameter vectors. A function $g$ of $\bm{x} \in \bm{\mathcal{X}}$, parameterized by $\bm\theta$ is written as $g(\bm{x};\bm\theta)$, when
the dependence on $\bm\theta$ is to be made explicit.

\subsection{Summary}
The remaining sections of the paper are organized as follows. Section \ref{section:probmodel} reviews normalizing flows, which are the central building block of the work herein presented. Section \ref{section:vmonf} introduces the proposed approach, \textit{variational mixtures of normalizing flows} (VMoNF), and the corresponding learning algorithm. Finally, experiments are presented in Section \ref{section:experiments}, and Section \ref{section:conclusions} concludes the paper with a brief discussion  and some pointers for future work.

\section{Normalizing Flows}
\label{section:probmodel}

\subsection{Introduction}
The best-known and most studied probability distributions, which are analitically
manageable, are rarely expressive enough for real-world complex datasets, such
as images or signals. However, they have properties that make them amenable to
work with, for instance, they allow for tractable parameter estimation,
they have closed-form likelihood functions, and sampling from them is simple.

One way to obtain more expressive models is to assume the existence of latent variables, leverage certain factorization
structures, and use well-known distributions for the individual factors of the product that
constitutes the model's joint distribution. By using these structures and
specific, well-chosen combinations of distributions (namely, conjugate prior-likelihood pairs),
these models are able to remain tractable - normally via bespoke estimation/inference/learning
algorithms.

Another approach to obtaining expressive probabilistic models is to apply
transformations to a simple distribution, and use the \emph{change of variables}
formula to compute probabilities in the transformed space. This is the basis
of \emph{normalizing flows}, an approach proposed by \textcite{shakir_nf},
and which has since evolved and developed into the basis of multiple state-of-the-art
techniques for density modelling and estimation \autocite{Glow}, \autocite{real-nvp}, \autocite{bnaf19},
\autocite{maf}.

\subsection{Change of Variables}
\label{cov}
Given a random variable $\bm{z} \in \mathbb{R}^D$, with probability density function $f_Z$,
and a bijective and continuous function $g(\: . \: ;\bm\theta)\ : \ \mathbb{R}^D\rightarrow \mathbb{R}^D$,
the probability density function $f_X$ of the random variable $\bm{x} = g(\bm{z})$ is given by
\begin{align}
    f_X(\bm{x}) &= f_Z(g^{-1}(\bm{x};\bm\theta))\Big|\det\Big(\frac{d}{d\bm{x}}g^{-1}(\bm{x};\bm\theta)\Big)\Big| \\
    &= f_Z(g^{-1}(\bm{x};\bm\theta))\Big|\det\Big(\frac{d}{d\bm{z}}g(\bm{z};\bm\theta) \bigg{|}_{\bm{z} = g^{-1}(\bm{x};\bm\theta)}\Big)\Big|^{-1},
\end{align} where $\det\Big(\frac{d}{d\bm{x}}g^{-1}(\bm{x};\bm\theta)\Big)$
is the determinant of the Jacobian matrix of $g^{-1}(\: . \: ;\bm\theta)$,
computed at $\bm{x}$.
Since $g(\: . \: ;\bm\theta)$ is a transformation parameterized by a parameter
vector $\bm\theta$, this expression can be optimized w.r.t. $\bm\theta$, with the
goal of making it approximate some arbitrary distribution. For this to be feasible,
the following have to be easily computable:
\begin{itemize}
    \item $f_Z$ - the starting probability density function
        (also called \emph{base density}). It is assumed that it has a closed-form
        expression. In practice, this is typically one of the basic distributions
        (Gaussian, Uniform, etc.)
    \item $\det\Big(\frac{d}{d\bm{x}}g^{-1}(\bm{x};\bm\theta)\Big)$ - the determinant
        of the Jacobian matrix of $g^{-1}$; for most transformations, this is not
        \q{cheap} to compute.
    \item The gradient of $\det\Big(\frac{d}{d\bm{x}}g^{-1}(\bm{x};\bm\theta)\Big)$
        w.r.t. $\bm\theta$; this is crucial for gradient-based optimization of
        $\bm\theta$ to be feasible. For most cases, this is not easily computable.
\end{itemize}
As will become clear, the crux of the \emph{normalizing flows} framework is to find
transformations that are expressive enough, and for which the determinants of their
Jacobian matrices, as well as the gradients of those determinants are both \q{cheap}
to compute.

\subsection{Normalizing Flows}
Consider $L$ transformations $h_\ell$, for $\ell = 0, 1, ..., L-1$ that fulfill the
three requirements listed above. Let each of those transformations be parameterizable
by a parameter vector $\bm\theta_\ell$, for $\ell = 0, 1, ..., L-1$. The
dependence on the parameter vectors will be implicit from here on.
Let $\bm{z_\ell} = h_{\ell-1} \circ h_{\ell-2} \circ ... \circ h_0(\bm{z_0})$, where
$\bm{z_0}$ is sampled from $f_Z$, the base density. Notice that, with this notation,
$\bm{z_L} = \bm{x}$. Furthermore, let $g$ be the composition of the $L$ transformations.
Applying the change of variables formula to
\begin{align}
    \bm{z_0} &\sim f_Z \\
    \bm{x} &= h_{L-1} \circ h_{L-2} \circ ... \circ h_0(\bm{z_0}),
\end{align}
noting that $g^{-1} = h^{-1}_0 \circ h^{-1}_1 \circ ... \circ h^{-1}_{L-1}$ and
using the chain rule for derivatives, leads to
\begin{align}
    f_X(\bm{x}) &= f_Z(g^{-1}(\bm{x}))\Big|\det\Big(\frac{d}{d\bm{x}}g^{-1}(\bm{x})\Big)\Big| \\
                        &= f_Z(g^{-1}(\bm{x}))\prod_{\ell=0}^{L-1}\Big|\det\Big(\frac{d}{d\bm{z_{\ell+1}}}h_{\ell}^{-1}(\bm{z_{\ell+1}})\Big)\Big| \\
                        &= f_Z(g^{-1}(\bm{x}))\prod_{\ell=0}^{L-1}\Big|\det\Big(\frac{d}{d\bm{x_{\ell}}}h_{\ell}(\bm{x_\ell})\bigg{|}_{\bm{x_\ell} = h_{\ell}^{-1}(\bm{z_{\ell+1}})}   \Big)\Big|^{-1} \label{eq:nflowderivation}
\end{align}
Replacing $h_{\ell}^{-1}(\bm{z_{\ell+1}}) = \bm{z_\ell}$ in (\ref{eq:nflowderivation}) leads to
\begin{align}
         f_X(\bm{x}) = f_Z(g^{-1}(\bm{x}))\prod_{\ell=0}^{L-1}\Big|\det\Big(\frac{d}{d\bm{z_{\ell}}}h_{\ell}(\bm{z_\ell})\Big)\Big|^{-1};
\end{align} taking the logarithm,
\begin{align}
    \log f_X(\bm{x}) = \log f_Z(g^{-1}(\bm{x})) - \sum_{\ell=0}^{L-1} \log \Big|\det\Big(\frac{d}{d\bm{z_{\ell}}}h_{\ell}(\bm{z_\ell})\Big) \Big|. \label{eq:nflowsfinal}
\end{align}
Depending on the task, one might prefer to replace the second term in (\ref{eq:nflowsfinal})
with a sum of log-absolute-determinants of the Jacobians of the inverse transformations.
This choice would imply replacing the minus sign before the sum with a plus sign:
\begin{align}
    &\log f_X(\bm{x}) = \nonumber \\
    &= \log f_Z(g^{-1}(\bm{x})) + \sum_{\ell=0}^{L-1} \log \Big|\det\Big(\frac{d}{d\bm{z_{\ell+1}}}h_{\ell}^{-1}(\bm{z_{\ell+1}})\Big) \Big|.
\end{align}
We started by assuming that the transformations $h_\ell$ fulfill the requirements
listed in Section \ref{cov}. For that reason, it is clear that the above expression
is a feasible objective for gradient-based optimization. In practice, this is carried
out by leveraging modern automatic differentiation and optimization frameworks \autocites{flowpp, Glow, real-nvp}.
Sampling from the resulting distribution is simply achieved by sampling from the base
distribution and applying the chain of transformations. Because of this, normalizing
flows can be used as flexible variational posteriors, in variational
inference settings, as well as density estimators.

\subsection{Examples of transformations}
\subsubsection{Affine Transformation}
An affine transformation is arguably the simplest choice; it can
stretch, sheer, shrink, rotate, and translate the space. It is simply achieved
by the multiplication by a matrix $A$ and summation of a bias vector $\bm{b}$:
\begin{align}
    \bm{z} &\sim p(\bm{z}) \\
    \bm{x} &= A\bm{z} + \bm{b}.
\end{align}
The determinant of the Jacobian of this transformation is simply the determinant
of $A$. However, in general, computing the determinant of a $D \times D$
matrix has $\mathcal{O}(D^3)$ computational complexity. For that reason, it is
common to use matrices with a certain structure that makes their determinants
easier to compute. For instance, if $A$ is triangular, its determinant is
the product of its diagonal's elements. The downside of using matrices that are
constrained to a certain structure is that they correspond to less flexible transformations.

It is possible, however, to design affine transformations whose Jacobian determinants
are of $\mathcal{O}(D)$ complexity and that are more expressive than simple
triangular matrices. \textcite{Glow} propose one such transformation. It
constrains the matrix $A$ to be decomposable as ${A = PL\big(U + \mbox{diag}(\bm{s})\big)}$,
where $\mbox{diag}(\bm{s})$ is a diagonal matrix whose diagonal's elements are
the components of vector $\bm{s}$. The following additional constrains are in place:
\begin{itemize}
    \item $P$ is a permutation matrix
    \item $L$ is a lower triangular matrix, with ones in the diagonal
    \item $U$ is an upper triangular matrix, with zeros in the diagonal
\end{itemize}
Given these constraints, the determinant of matrix $A$ is simply the product
of the elements of $\bm{s}$.

\begin{figure}[!htb]
  \begin{subfigmatrix}{2}
    \subfigure[]{\includegraphics[width=0.49\linewidth]{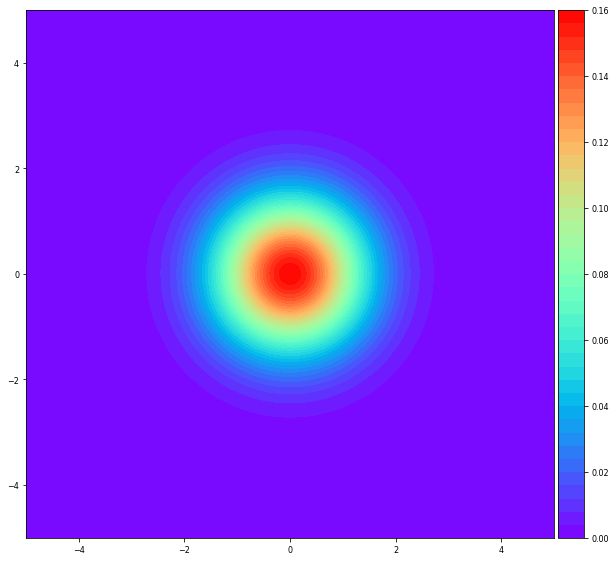}}
    \subfigure[]{\includegraphics[width=0.49\linewidth]{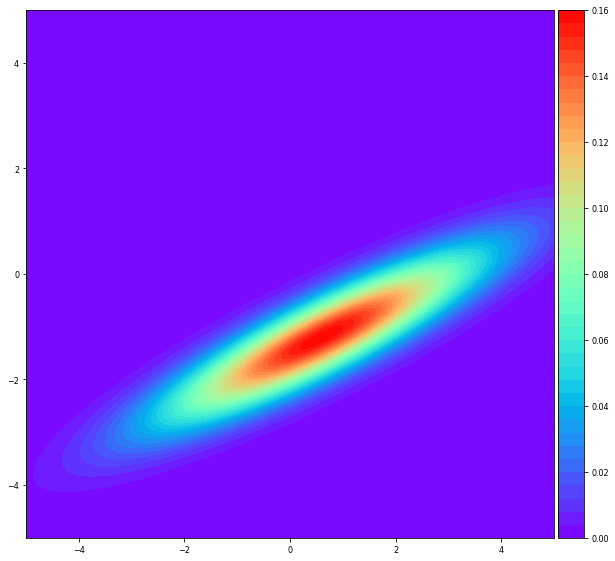}}
  \end{subfigmatrix}
    \caption{(a) Density of a Gaussian distribution with $\mu = [0, 0]$ and $\Sigma = I$
    (b) Density of the distribution that results from applying some affine transformation to
    the Gaussian distribution in (a)
    }
  \label{fig:affine}
\end{figure}

\subsubsection{PReLU Transformation}
Intuitively, introducing non-linearities endows normalizing flows with more flexibility to
represent complex distributions. This can be done in a similar fashion to the
activation functions used in neural networks. One example of that is the parameterized
rectified linear unit (PReLU) transformation. It is defined in the following manner, for
a $D$-dimensional input:
\begin{align}
f(\bm{z}) = [f_1(z_1), f_2(z_2), ..., f_D(z_D)],
\end{align} where
\begin{align}
f_i(z_i) =
    \begin{cases}
        z_i,              & \text{if } z_i\geq 0, \\
        \alpha z_i,       & \text{otherwise}.
    \end{cases}
\end{align}
In order for the transformation to be invertible, it is necessary
that $\alpha > 0$.
Let us define a function $j(.)$ as
\begin{align}
j(z_i) =
    \begin{cases}
       1 ,              & \text{if } z_i \geq 0, \\
       \alpha ,       & \text{otherwise};
    \end{cases}
\end{align}
it is trivial to see that the Jacobian of the transformation is a diagonal
matrix, whose diagonal elements are $j(z_i)$:
\begin{align}
  J(f(z)) =
  \begin{bmatrix}
      j(z_1) & & & \\
      & j(z_2) & & \\
      & & \ddots & \\
      & & & j(z_D)
  \end{bmatrix}.
\end{align}
With that in hand, it is easy to arrive at the log-absolute-determinant of this transformation's
Jacobian, which is given by $\sum_{i=1}^D \log \big| j(z_i) \big|$

\begin{figure}[!htb]
  \begin{subfigmatrix}{2}
    \subfigure[]{\includegraphics[width=0.49\linewidth]{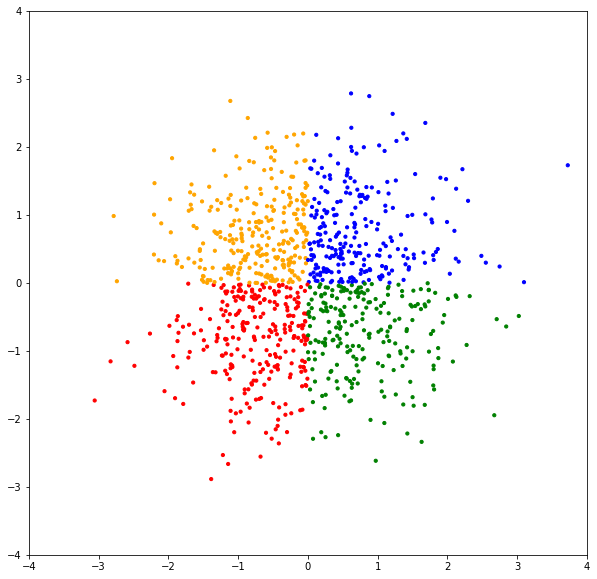}}
    \subfigure[]{\includegraphics[width=0.49\linewidth]{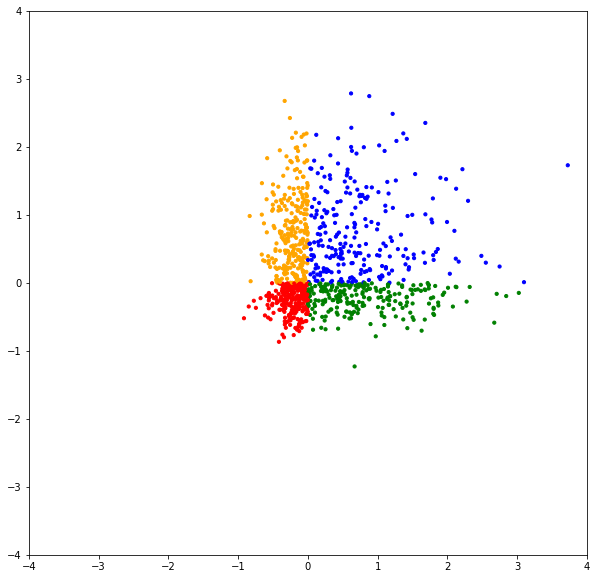}}
  \end{subfigmatrix}
    \caption{(a) Samples from of a Gaussian distribution with $\mu = [0, 0]$ and $\Sigma = I$.
    The samples are colored according to the quadrant they belong to. (b) Samples from the
    distribuion in a) transformed by a PReLU transformation.}
  \label{fig:prelu}
\end{figure}

\begin{figure}[!htb]
  \begin{subfigmatrix}{3}
    \subfigure[]{\includegraphics[width=0.31\linewidth]{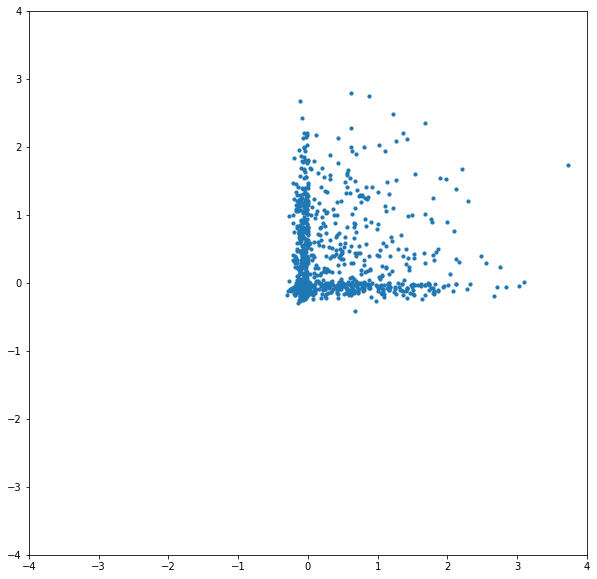}}
    \subfigure[]{\includegraphics[width=0.31\linewidth]{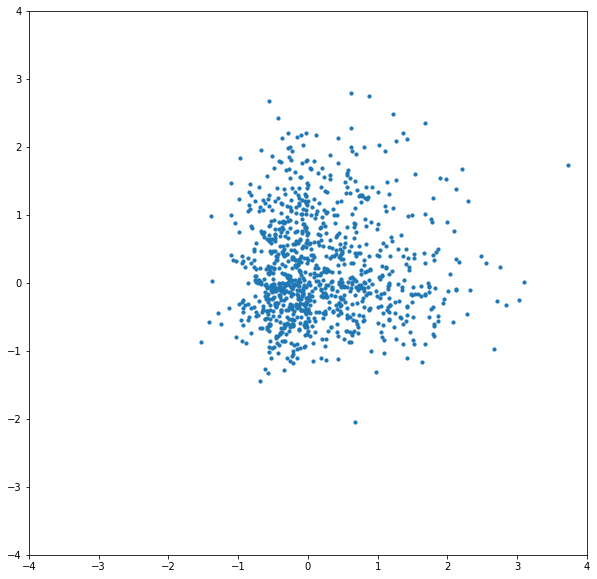}}
    \subfigure[]{\includegraphics[width=0.31\linewidth]{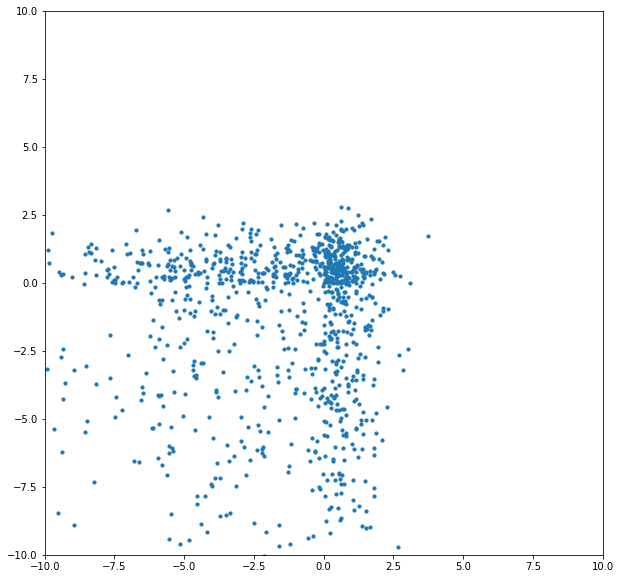}}
  \end{subfigmatrix}
    \caption{Samples from a Gaussian with $\mu = [0, 0]$ and $\Sigma = I$, transformed
    by PReLU transformations with different $\alpha$ parameters. (a) $\alpha = 0.1$
    (b) $\alpha = 0.5$ (c) $\alpha = 5$}
  \label{fig:prelu2}
\end{figure}

\subsubsection{Batch-Normalization Transformation}
\textcite{real-nvp} propose a batch-normalization transformation, similar to
the well-known batch-normalization layer normally used in neural networks. This
transform simply applies a rescaling, given the batch mean $\tilde\mu$ and variance
${\tilde\sigma}^2$:
\begin{align}
    f(z) = \frac{z - \tilde\mu}{\sqrt{{\tilde\sigma}^2 + \epsilon}},
\end{align} where $\epsilon \ll 1$ is a term used to ensure that there never is
a division by zero. This transformation's Jacobian is trivial: 
\begin{align}
    \prod_{i=1}^D \frac{1}{\sqrt{{\tilde\sigma}_i^2 + \epsilon}}.
\end{align}

\subsubsection{Affine Coupling Transformation}
As mentioned previously, one of the active research challenges within the
normalizing flows framework is the search and design of transformations that
are sufficiently expressive and whose Jacobians are not computationally heavy. One brilliant
example of such transformations, proposed by \textcite{real-nvp}, is
called affine coupling layer.

This transformation is characterized by two arbitrary functions $s(.)$ and
$t(.)$, as well as a mask that splits an input $\bm{z}$ of dimension $D$ into
two parts, $\bm{z_1}$ and $\bm{z_2}$. In practice, $s(.)$ and $t(.)$ are neural
networks, whose parameters are to be optimized so as to make the transformation
approximate the desired output distribution. The outputs of $s(.)$ and $t(.)$
need to have the same dimension as $\bm{z_1}$. This should be taken into account when
designing the mask and the functions $s(.)$ and $t(.)$. The transformation is defined as:
\begin{align}
    \begin{cases}
    \bm{x_1} &= \bm{z_1} \odot \exp\big(s(\bm{z_2})\big) + t(\bm{z_2}) \\
    \bm{x_2} &= \bm{z_2}.
    \end{cases}
\end{align}
To see why this transformation is suitable to being used within the framework
of normalizing flows, let us derive its Jacobian.
\begin{itemize}
    \item $\frac{\partial \bm{x_2}}{\partial \bm{z_2}} = I$, because $\bm{x_2} = \bm{z_2}$.
    \item $\frac{\partial \bm{x_2}}{\partial \bm{z_1}}$ is a matrix of zeros, because $\bm{x_2}$ does not depend on $\bm{z_1}$.
    \item $\frac{\partial \bm{x_1}}{\partial \bm{z_1}}$ is a diagonal matrix,
        whose diagonal is simply given by $\exp\big(s(\bm{z_2})\big)$, since those values are
        constant w.r.t $\bm{z_1}$ and they are multiplying each element of $\bm{z_1}$.
    \item $\frac{\partial \bm{x_1}}{\partial \bm{z_2}}$ is not needed,
        as will become clear ahead.
\end{itemize}

Writing the above in matrix form:

\begin{align}
    J_{f(z)} &=
        \begin{tikzpicture}[decoration=brace, baseline=-\the\dimexpr\fontdimen22\textfont2\relax ]
            \matrix (m) [matrix of math nodes,left delimiter=[,right delimiter={]}, ampersand replacement=\&] {
                \mbox{\Large$\frac{\partial \bm{x_1}}{\partial \bm{z_1}}$} \& \mbox{\Large$\frac{\partial \bm{x_1}}{\partial \bm{z_2}}$} \\
                \mbox{\Large$\frac{\partial \bm{x_2}}{\partial \bm{z_1}}$} \& \mbox{\Large$\frac{\partial \bm{x_2}}{\partial \bm{z_2}}$} \\
            };
        \end{tikzpicture} \\
    &=
        \begin{tikzpicture}[decoration=brace, baseline=-\the\dimexpr\fontdimen22\textfont2\relax ]
            \matrix (m) [matrix of math nodes,left delimiter=[,right delimiter={]}, ampersand replacement=\&] {
                \mbox{diag}\Big(\exp\big(s(\bm{z_2})\big)\Big) \& \mbox{\Large$\frac{\partial \bm{x_1}}{\partial \bm{z_2}}$} \\
                \mbox{\Large$\bm{0}$} \& \mbox{\Large$I$} \\
            };
        \end{tikzpicture}
\end{align} shows that the Jacobian matrix is (upper) triangular. Its determinant - the
only thing we need, in fact - is therefore easy to compute: it is simply the
product of the diagonal elements. Moreover, part of the diagonal is simply
composed of ones. The determinant, and the log-absolute-determinant become
\begin{align}
    \det\big(J_{f(z)}\big) &= \prod_i \exp\big(s(\bm{z_2}^{(i)})\big) \\
    \log \Big|\det\big(J_{f(z)}\big)\Big| &= \sum_i s(\bm{z_2}^{(i)}),
\end{align} where $\bm{z_2^{(i)}}$ is the $i$-th element of $\bm{z_2}$.
Since a single affine coupling layer does not transform all of the elements in
$\bm{z}$, in practice several layers are composed, and each layer's mask is changed
so as to make all dimensions affect each other. This can be done, for instance, with
a checkerboard pattern, which alternates for each layer. In the case of image inputs,
the masks can operate at the channel level.

\subsubsection{Masked Autoregressive Flows}
Another ingenious architecture for normalizing flows has been proposed by \textcite{maf}.
It is called masked autoregressive flow (MAF). Let $\bm{z}$ be a sample from
some base distribution, with dimension $D$. MAF transforms $\bm{z}$ into an
observation $\bm{x}$, of the same dimension, in the following manner:
\begin{align}
x_i = z_i \exp(\alpha_i) + \mu_i \\
(\mu_i, \alpha_i) = g(\bm{x_{1:i-1}}).
\end{align}
In the above expression $g$ is some arbitrary function. The inverse transform of
MAF is trivial, because, like the affine coupling layer, MAF uses $g$ to parameterize
a shift, $\mu$, and a log-scale, $\alpha$, which translates to the fact that the
function $g$ itself does not need to be inverted:
\begin{align}
z_i = (x_i - \mu_i)\exp(-\alpha_i).
\end{align}
Moreover, the autoregressive structure of the transformation constrains the
Jacobian to be triangular, which renders the determinant effortless to compute: 
\begin{align}
\det\big( J_{f(\bm{z})} \big) &= \prod_{i=1}^{D} \exp(\alpha_i), \\
\log \Big| \det \big( J_{f(\bm{z})} \big) \Big| &= \sum_{i=1}^{D} \alpha_i.
\end{align}
As stated above, the function $g$ used to obtain $\mu_i$ and $\alpha_i$ can be
arbitrary. However, in the original paper, the function proposed a masked
autoencoder for distribution estimation (MADE), as described by \textcite{MADE}.

Much like the partitioning in the affine coupling layer, the assumption of
autoregressiveness (and the ordering of the elements of $\bm{x}$
for which that assumption is held) carries an inductive bias with it. Again,
like with the affine coupling layer, this effect is minimized in practice by
stacking layers with different element orderings.

\subsection{Fitting Normalizing Flows}

Generally speaking, normalizing flows can be used in one of two scenarios:
(direct) density estimation, where the goal is to optimize the parameters
so as to make the model approximate the distribution of some observed set of data;
in a variational inference scenario, as way of having a flexible variational
posterior. The second scenario is out of the scope of this work.

The task of density estimation with normalizing flows reduces to finding the
optimal parameters of a parametric model. In general, there are two ways to go about estimating
the parameters of a parametric model,
given data: MLE and MAP. In the case of normalizing flows, MLE is the usual
approach\footnote{In theory it is possible to place a prior on the normalizing
flow's parameters and do MAP estimation. To accomplish this, similar strategies
to those used in Bayesian Neural Networks would have to be used.}. To fit a normalizing
flow via MLE, a gradient based optimizer is used to minimize
$\hat{\mathcal{L}}(\bm\theta) = - \mathbb{E}[\log p(\bm{x}|\bm\theta)]$.
However, this expectation is generally not accesible, since we have only
finite samples of $\bm{x}$. Because of that, the parameters are estimated
by optimizing an approximation of that expectation: $ - \frac{1}{N} \sum_{i=1}^N \log p(\bm{x_i} | \bm\theta)$.

To perform optimization on this objective, stochastic gradient descent (SGD) - and
its variants -  is the most commonly used algorithm. In general terms, SGD is an
approximation of gradient descent, which rather than using the actual gradient,
at time step $t$, to update the variables under optimization, works by computing
several estimates of that gradient and using those estimates instead. This is
done by partitioning the data in mini-batches, and computing the loss function
and respective gradients over those mini-batches. This way, one pass through
data - an \emph{epoch} - results in several parameter updates.

\section{Variational Mixture of Normalizing Flows}
\label{section:vmonf}

\subsection{Introduction}
\label{subsection:vmonf-intro}

The ability of leveraging domain knowledge to endow a probabilistic model with
structure is often useful. The goal of this work is to devise a model that combines
the flexibility of normalizing flows with the ability to exploit class-membership
structure. This is achieved by learning a mixture of normalizing flows, via
optimization of a variational objective, for which the variational posterior
over the class-indexing latent variables is parameterized by a neural network.
Intuitively, this neural network should learn to place similar instances of
data in the same class, allowing each component of the mixture to be fitted
to a cluster of data.

\subsection{Model Definition}

Let us define a mixture model, where each of
the $K$ components is a density parameterized by a normalizing flow. For simplicity,
consider that all of the $K$ normalizing flows have the same
architecture\footnote{This is not a requirement,
and in cases where we have classes with different levels of complexity, we can
have components with different architectures. However, the training procedure
does not guarantee that the most flexible normalizing flow is "allocated"
to the most complex cluster. This is an interesting direction for future
research.}, i.e., they are all composed of the same stack of transformations,
but they each have their own parameters.

Additionally, let $q(z|\bm{x};\gamma)$ be a neural network with a $K$-class softmax
output, with parameters $\bm\gamma$. This network will receive as input an instance from the
data, and produce the probability of that instance belonging to each of the
$K$ classes.

Recall the evidence lower bound (the dependence of $q$ on $x$ is made explicit):
\begin{equation*}
    \text{ELBO} = \mathbb{E}_q [\log p(\bm{x}, z)] - \mathbb{E}_q [\log q(z|\bm{x})].
\end{equation*}

Let us rearrange it:
\begin{align}
    \text{ELBO} &= \mathbb{E}_q [\log p(\bm{x}|z)] + \mathbb{E}_q [\log p(z)] - \mathbb{E}_q [\log q(z|\bm{x})]
        \label{eq:threepartelbo} \\
    &= \mathbb{E}_q [\log p(\bm{x}|z) + \log p(z) - \log q(z|\bm{x})] \label{eq:simplerelbo}
\end{align}

Since $q(z|\bm{x})$ is given by the forward-pass of a neural network, and is therefore
straightforward to obtain, the expectation in (\ref{eq:simplerelbo}) is given by
computing the expression inside the expectation for each possible value of $z$,
and summing the obtained values, weighed by the probabilities given by the variational posterior:
\begin{align}
    \text{ELBO} = \sum_{z=1}^K q(z|\bm{x})\big(\log p(\bm{x}|z) + \log p(z) - \log q(z|\bm{x})\big).
\end{align}
Thus, the whole ELBO is easy to compute, provided that each of the terms inside
the expectation is itself easy to compute. Let us consider each of those terms:
\begin{itemize}
    \item $\log p(\bm{x}|z)$ is the log-likelihood of $\bm{x}$ under the normalizing
        flow indexed by $z$. It was shown in the previous section how to compute
        this.
    \item $\log p(z)$ is the log-prior of the component weights. For simplicity,
        let us assume this is set by the modeller. When nothing is known about
        the component weights, the best assumption is that they are uniform.
    \item $- \log q(z|\bm{x})$ is the negative logarithm of the output of the encoder.
\end{itemize}

\begin{figure}[!htb]
  \centering
  \includegraphics[width=0.5\linewidth]{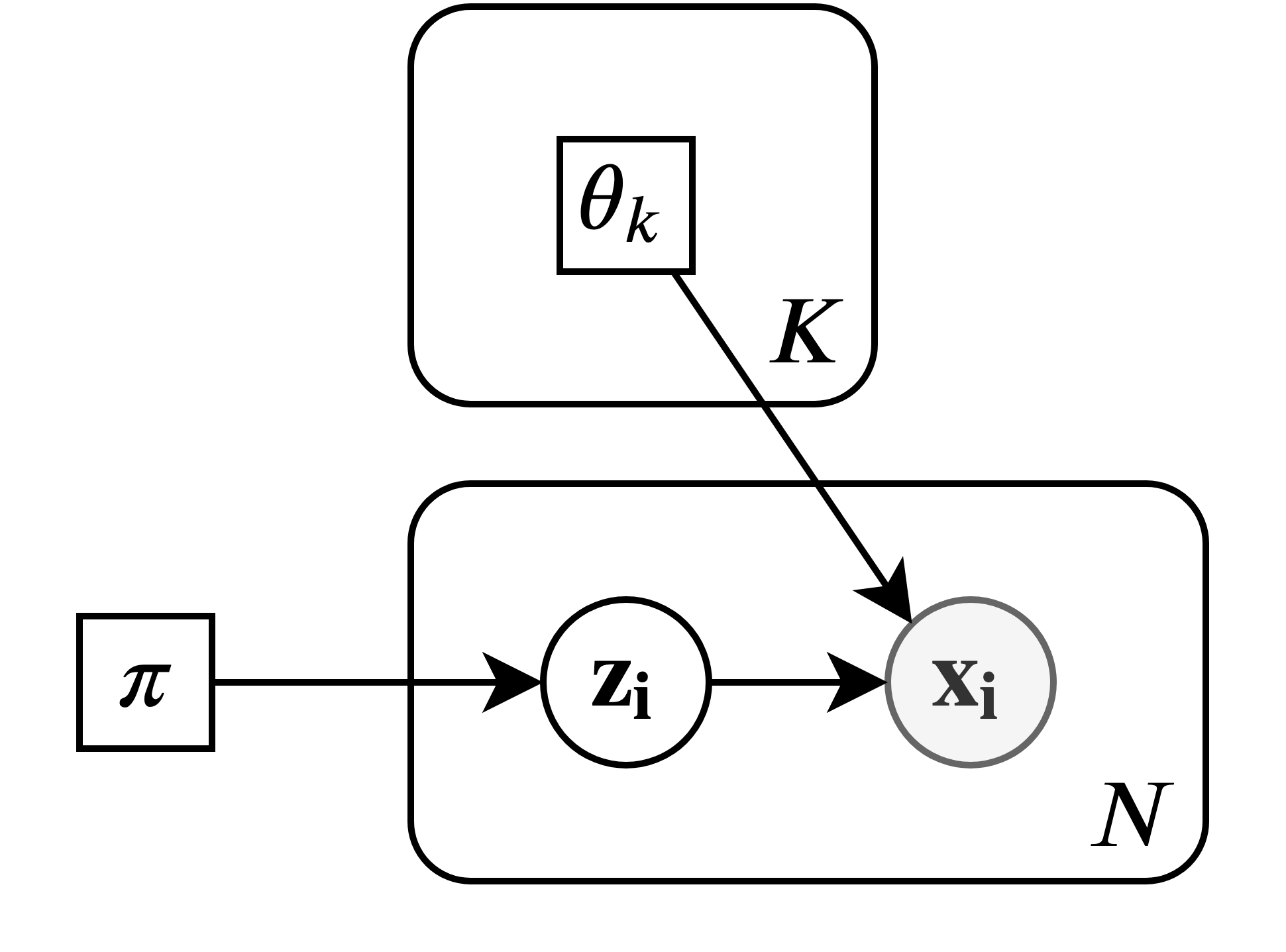}
  \caption{Plate diagram of a mixture of $K$ normalizing flows. $\bm\theta_k$ is the
    parameter vector of component $k$.}
  \label{fig:plate}
\end{figure}

\begin{figure}[!htb]
  \centering
  \includegraphics[width=0.85\linewidth]{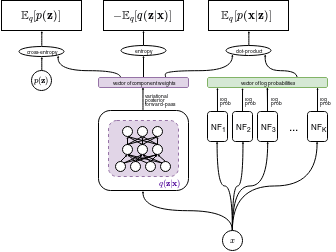}
  \caption{Overview of the training procedure of the VMoNF.}
  \label{fig:modeloverview}
\end{figure}

Let us call this model \emph{variational mixture of normalizing flows} (VMoNF). For an overview of
the model, consider Figures \ref{fig:plate} and \ref{fig:modeloverview}

In a similar fashion to the variational auto-encoder, proposed by \textcite{vaepaper},
a VMoNF is fitted by jointly optimizing the parameters of the variational
posterior $q(z|\bm{x}; \bm\gamma)$ and the parameters of the generative process
$p(\bm{x}|z; \bm\theta)$.
After training, the variational posterior naturally induces a clustering on
the data, and can be directly used to assign new data points to the discovered clusters.
Moreover, each of the fitted components can be used to generate samples from the
cluster it \q{specialized} in.

\subsection{Implementation}

To implement and test the proposed model, Python was the chosen language. More
specifically, this work heavily relies on the PyTorch \autocite{pytorch} package
and framework for automatic differentiation. Moreover, the parameter optimization
is done via stochastic optimization, namely using the Adam optimizer, proposed by
\textcite{adam}.

Figure \ref{fig:modeloverview} gives an overview of the training procedure:
\begin{enumerate}
    \item The \emph{log-probabilities} given by each component of the mixture
    are computed.
    \item The values of the variational posterior probabilities for each
    component are computed.
    \item With the results of the previous steps, all three terms of the ELBO
    are computable.
    \item The ELBO and its gradients w.r.t the model parameters are computed
    and the parameters are updated.
    \item Steps 1 to 4 are repeated until some stopping criterion is met.
\end{enumerate}

\section{Experiments}
\label{section:experiments}

In this section, the proposed model is applied to two benchmark synthetic datasets (Pinwheel
and Two-circles) and one real-world dataset (MNIST). On one of the synthetic
datasets, one shortcoming of the model is brought to attention, but is overcome
in a semi-supervised setting. On the real-world dataset, the model's clustering
capabilities are evaluted, as well as its capacity to model complex distributions.

A technique inspired in the work of \textcite{mixae} was employed to improve training
speed and quality of results. This consists in dividing the inputs of the softmax layer in
the variational posterior by a \q{temperature} value, $T$, which follows
an exponential decay schedule during training. Intuitively, this makes the
variational posterior \q{more certain} as training proceeds, while allowing all
components to be generally exposed to the whole data, during the initial epochs.
This discourages components from being \q{subtrained} during the initial epochs
and, subsequently, from being prematurely discarded by the variational posterior.

\subsection{Toy datasets}
\subsubsection{Pinwheel dataset}

\begin{figure}[!htb]
  \begin{subfigmatrix}{2}
    \subfigure[]{\includegraphics[width=0.49\linewidth]{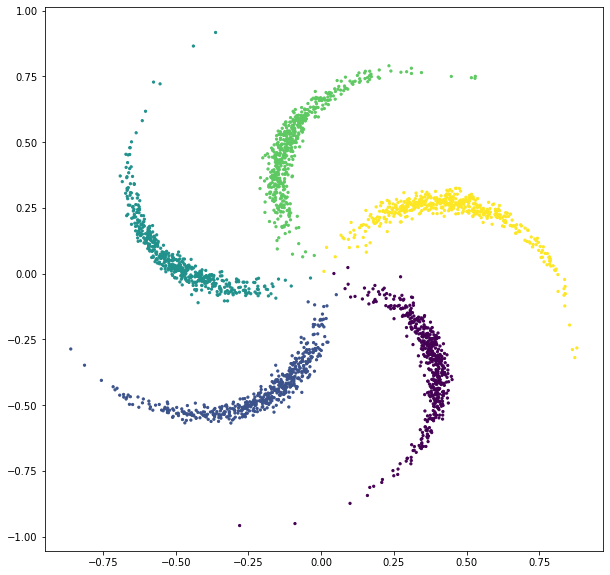}}
    \subfigure[]{\includegraphics[width=0.49\linewidth]{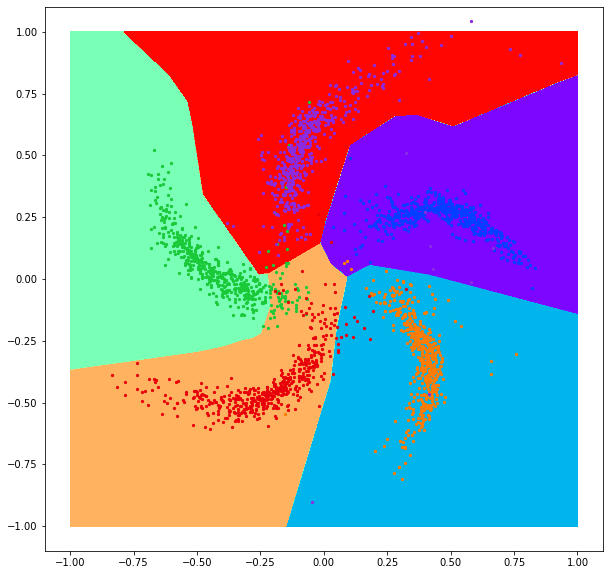}}
  \end{subfigmatrix}
    \caption{(a) Original dataset. (b) Samples from the learned model. Each
dot is colored according to the component it was sampled from. The background
colors denote the regions where each component has maximum probability assigned
by the variational posterior. (Note that the background colors were chosen
so as to not match the dot colors, otherwise the dots would not be visible)}
  \label{fig:pinwheel}
\end{figure}

This dataset is constituted by five non-linear \q{wings}. See Figure \ref{fig:pinwheel}
for the results of running the model on this dataset. As expected, the variational
posterior has learned to partition the space so as to attribute each \q{wing} to
a component of the mixture. This partitioning is imperfect in regions of space
that have low probability for every component.

This experiment consisted of training on 2560 data points (512 per class) using
the Adam optimizer, with a learning rate of 0.001, a
mini-batch size of 512, during 400 epochs. The variational posterior was parameterized
by a multi-layer perceptron, with 1 hidden layer of dimension 3, and a
softmax output. Each component of the mixture was a RealNVP with 8 blocks, each
block with multi-layer perceptrons, with 1 hidden layer of dimension 8 as the
$s(.)$ and $t(.)$ functions of the affine coupling layers.

\subsubsection{Two-circles dataset}
\begin{figure}[!htb]
  \begin{subfigmatrix}{2}
    \subfigure[]{\includegraphics[width=0.49\linewidth]{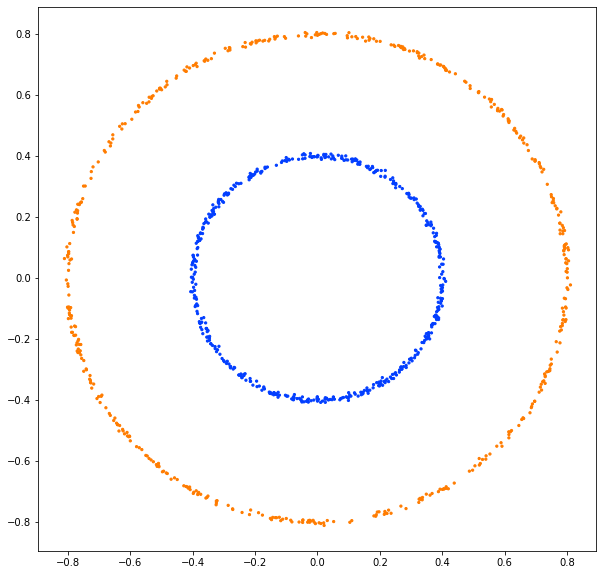}}
    \subfigure[]{\includegraphics[width=0.49\linewidth]{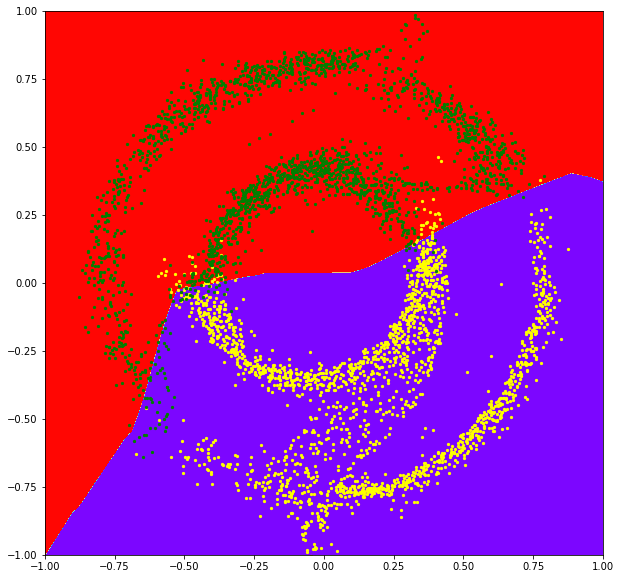}}
  \end{subfigmatrix}
    \caption{(a) Original dataset. (b) Samples from the learned model, without
    any labels. Coloring logic is the same as in \ref{fig:pinwheel}.}
\label{fig:twocircles}
\end{figure}

\begin{figure}[!htb]
  \begin{subfigmatrix}{2}
    \subfigure[]{\includegraphics[width=0.49\linewidth]{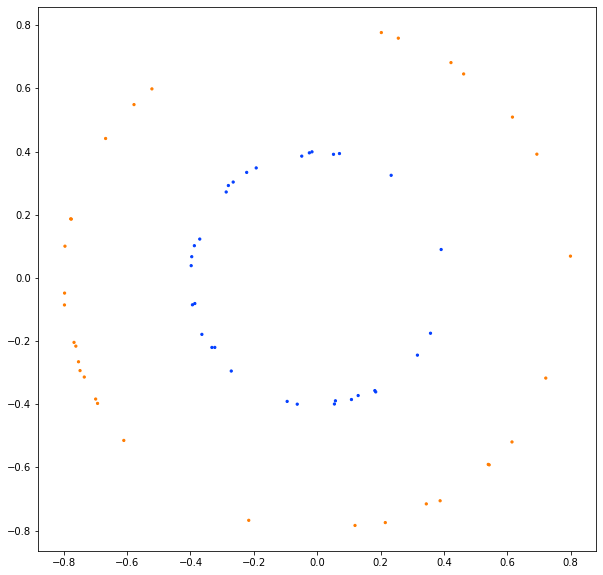}}
    \subfigure[]{\includegraphics[width=0.49\linewidth]{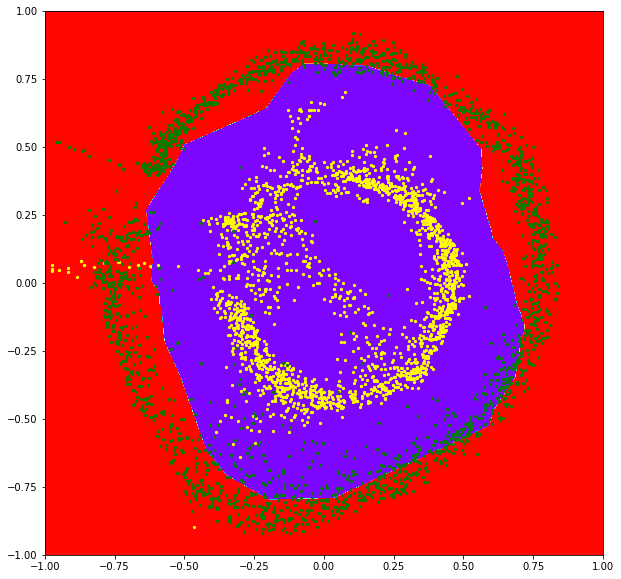}}
  \end{subfigmatrix}
    \caption{(a) Labeled points used in semi-supervised scenario. (b) Samples
    from the model trained in the semi-supervised scenario.}
\label{fig:twocircles-semisup}
\end{figure}

This dataset consists of two concentric circles. The experiment on this dataset,
shown on Figure \ref{fig:twocircles}, makes evident one shortcoming of this
model: the way in which the variational posterior partitions the space is
not necessarily guided by the intrisic structure in the data. In the case of
the two-circles dataset, it was found that the most common space partitioning
induced by the model consisted simply of splitting into two half-spaces. However, in
a semi-supervised setting, this behaviour can be corrected and the model
successfully learns to separate the two circles, as shown in Figure
\ref{fig:twocircles-semisup}. In this setting, the model was pretrained on
the labeled instances for some epochs and then trained with the normal procedure.
In the semi-supervised setting, the model has the chance to refine both the
variational posterior and each of the components, thus making better use of
the unlabeled data in the unsupervised phase of the training. As is clearly
visible in Figure \ref{fig:twocircles-semisup}, the model struggles with
learning full, closed, circles; this is because it is unable to \q{pierce a hole}
in the base distribution, due to the nature of the transformations that are
applicable. Thus, to model a circle, the model has to learn to stretch the blob
formed by the base distribution, and \q{bend it over itself}. This difficulty
is also what keeps the model from learning a structurally interesting solution
in the fully unsupervised case: it is easier to learn to distort space so as to
learn a multimodal distribution that models half of the two circles. Moreover,
the points in diametrically opposed regions of the same circle are more dissimilar
(in the geometrical sense) than points in the same region of the two circles.
Therefore, when completely uninformed by labels, the variational posterior's
layers will tend to have similar activations for points in the latter case, and
thus tend to place them in the same class.

The unsupervised learning experiment consisted of training on 1024 datapoints,
512 per class; using the Adam optimizer, with a learning rate
of 0.001, a mini-batch size of 128, during 500 epochs.
The semi-supervised learning experiment consisted of training on 1024 unlabeled
datapoints, 512 per class and 64 labeled data points, 32 per class. The model
was first pretrained during 300 epochs solely on the 32 labeled data points, using
the labels to selectively optimize each component of the mixture, as well as
to optimize the variational posterior by minimizing a binary cross-entropy loss.
After pretraining, the model was trained by interweaving supervised epochs - like
in pretraining - with unsupervised epochs. Optimization was carried out using the
Adam optimizer, with a learning rate of 0.001, a mini-batch size of 128, during 500 epochs.
For both the unsupervised and the semi-supervised experiments, the neural network
used to parameterize the variational posterior was a multi-layer perceptron, with
2 hidden layers of dimension 16, and with a softmax output. Each component of the
mixture was a RealNVP with 10 blocks, each block with multi-layer perceptrons,
with 1 hidden layer of dimension 8, as the $s(.)$ and $t(.)$ functions of the
affine coupling layers.

\subsection{Real-world dataset}
\begin{figure}[!htb]
  \centering
  \includegraphics[width=0.85\linewidth]{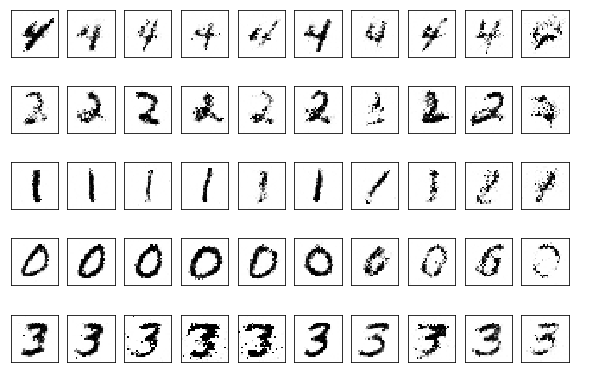}
  \caption{Samples from the fitted mixture components. Each row is sampled
  from the same component}
  \label{fig:mnist_samples}
\end{figure}

In this subsection, the proposed model is evaluated on the well-known MNIST
dataset \autocite{MNIST}. This dataset consists of images of handwritten digits.
The grids are of dimension 28 x 28 and were flattened to vectors of dimension 784
for training. For this experiment, only the images corresponding to the digits
from 0 to 4 were considered. The normalizing flow model used for the components
was a MAF, with 5 blocks, whose internal MADE layers had 1 hidden layer of dimension
200. The variational posterior was parameterized by a multi-layer perceptron, with
1 hidden layer of dimension 512. The model was trained for 100 epochs, with a mini-batch
size of 100. The Adam optimizer was used, with a learning rate of 0.0001, and with
a weight decay parameter of 0.000001. In Figure \ref{fig:mnist_samples}, samples
from the components obtained after training can be seen. Moreover, a normalized
contingency table is presented, where the performance of the variational posterior
as a clustering function can be assessed. Note that the cluster indices induced
by the model have no semantic meaning.

\begin{table}[ht]
\centering
\begin{tabular}{cccccc}
\toprule
\diagbox[trim=lr]{True\\label}{Cluster\\index} &         0 &         1 &         2 &         3 &         4 \\
\midrule
0    &  0.000602 &  0.012432 &  0.002807 &  0.982555 &  0.001604 \\
1    &  0.002139 &  0.020146 &  0.977001 &  0.000178 &  0.000535 \\
2    &  0.000802 &  0.952276 &  0.011630 &  0.007219 &  0.028073 \\
3    &  0.001558 &  0.479455 &  0.300682 &  0.004284 &  0.214021 \\
4    &  0.646166 &  0.347273 &  0.005125 &  0.001435 &  0.000000 \\
\bottomrule
\end{tabular}
\caption{Normalized contingency table for the clustering induced by the model}
\label{table:contingency}
\end{table}

From Table \ref{table:contingency} and Figure \ref{fig:mnist_samples} it is possible
to see that although there is some confusion, the model successfully clusters
the MNIST digits.

\section{Conclusions}
\label{section:conclusions}

\subsection{Conclusions}
\label{subsection:conclusions}
Deep generative modelling is an active research avenue that will keep being
developed and improved, since it lends itself to extremely useful applications,
like anomaly detection, synthetic data generation, and, generally speaking,
uncovering patterns in data.
Overall, the initial idea of the present work stands validated by the experiments - 
it is possible to learn mixtures of normalizing flows via the proposed procedure - as well
as by recently published similar work \autocites{RAD}{semisuplearning_nflows}.
The proposed method was tested on two synthetic datasets, succeeding with ease
on one of them, and struggling with the other one. However, when allowed to learn
from just a few labels, it was able to successfully fit the data it previously
failed on. On the real-world dataset, the model's clustering capability was tested,
as well as its ability to generate realistic samples, with some success.
During the experiments, it became evident that, similarly to what happens with
the majority of neural-network-based models, in order to successfully fit the
proposed model to complex data, some fine tuning is required, both in terms of the
training procedure, as well as in terms of the architecture of the blocks that
constitute the model. In the following subsection, some proposals and ideas for
future work and for tackling some of the observed shortcomings are proposed.

\subsection{Discussion and Future Work}
\label{subsection:future}

After the work presented here, some observations and future research questions
and ideas arise:
\begin{itemize}
    \item The main shortcoming of the proposed model, specially in its fully
    unsupervised variant, is that there is no way to incentivize the variational
    posterior to partition the space in the intuitively correct manner. Moreover, the
    variational posterior generally performs poorly in regions of space where there are few
    or no training points. This suggests that the model could benefit from a consistency
    loss regularization term. In fact, this idea has been pursued by \textcite{semisuplearning_nflows}.
    \item Some form of weight-sharing strategy between components is also an interesting
    point for future research. It is plausible that, this way, components could
    share \q{concepts} and latent representations of data, and use their non-shared
    weights to \q{specialize} in their particular cluster of data. Take,
    for instance, the Pinwheel dataset: in principle, the five normalizing flows
    could share a stack of layers that learned to model the concept of wing,
    each component then having a non-shared stack of blocks that would only
    need to model the correct rotation of its respective wing.
    \item During the experimentation phase, it was found that a balance between
    the complexity of the variational posterior and that of the components of
    the mixture, is crucial for the convergence to interesting solutions. This
    is intuitive: if the components are too complex, the variational posterior
    tends to ignore most of them and assigns most points to a single or few components.
    \item The fact that in some cases the variational posterior ignores components
    and \q{chooses} not to use them can hypothetically be exploited in the scenarios
    where the number of clusters is unknown. If the dynamics of what drives the
    variational posterior to ignore components can be understood, perhaps they
    can be actively tweaked (via architectural choices, training procedure and
    hyperparameters, for example) to benefit the modelling task in such a scenario.
    \item Related to the previous point, one first experiment could be to update
    the prior ($p(z)$) (for example, every epoch), based on the responsabilities
    given by the variational posterior.
    \item The effect of using different architectures for the neural networks used
    was not evaluated. It is likely, for instance, that convolutional architectures
    would produce better results in the real world dataset.
\end{itemize}

\printbibliography

\end{document}